\def\year{2021}\relax
\documentclass[letterpaper]{article} 
\usepackage{aaai21}  
\usepackage{times}  
\usepackage{helvet} 
\usepackage{courier}  
\usepackage[hyphens]{url}  
\usepackage{graphicx} 
\usepackage{natbib}  
\usepackage{caption} 
\usepackage{amsmath}
\usepackage{booktabs}
\usepackage{mathtools}
\usepackage{amssymb}
\usepackage{pifont}

\title{RESA: Recurrent Feature-Shift Aggregator for Lane Detection}
\author {
    Tu Zheng\textsuperscript{\rm 1,2}\thanks{Equal Contribution},
    Hao Fang\textsuperscript{\rm 1}\footnotemark[1],
    Yi Zhang\textsuperscript{\rm 1},
    Wenjian Tang\textsuperscript{\rm 2},
    Zheng Yang\textsuperscript{\rm 2},
    HaiFeng Liu\textsuperscript{\rm 1},
    Deng Cai\textsuperscript{\rm 1,2}\thanks{Deng Cai is the corresponding author.}\\
}
\affiliations {
    \textsuperscript{\rm 1} State Key Lab of CAD\&CG, College of Computer Science, Zhejiang University, China  \\
    \textsuperscript{\rm 2} Fabu Inc., Hangzhou, China \\
    zhengtuzju@gmail.com \quad  fanghao\_zju@163.com \quad  \{tangwenjian, yangzheng\}@fabu.ai \\ \quad \{yizhang2019, haifengliu, dcai\}@zju.edu.cn
}
\begin{document}

\maketitle

\begin{abstract}
Lane detection is one of the most important tasks in self-driving. Due to various complex scenarios~(\emph{e.g.}, severe occlusion, ambiguous lanes, \emph{etc}.) and the sparse supervisory signals inherent in lane annotations, lane detection task is still challenging. Thus, it is difficult for the ordinary convolutional neural network (CNN) to train in general scenes to catch subtle lane feature from the raw image. In this paper, we present a novel module named REcurrent Feature-Shift Aggregator (RESA) to enrich lane feature after preliminary feature extraction with an ordinary CNN. RESA takes advantage of strong shape priors of lanes and captures spatial relationships of pixels across rows and columns. It shifts sliced feature map recurrently in vertical and horizontal directions and enables each pixel to gather global information. RESA can conjecture lanes accurately in challenging scenarios with weak appearance clues by aggregating sliced feature map. Moreover, we propose a Bilateral Up-Sampling Decoder that combines coarse-grained and fine-detailed features in the up-sampling stage. It can recover the low-resolution feature map into pixel-wise prediction meticulously. Our method achieves state-of-the-art results on two popular lane detection benchmarks~(CULane and Tusimple). Code has been made available at: https://github.com/ZJULearning/resa. 
\end{abstract}

\section{Introduction}
Lane detection is an essential task in the computer vision community. It could serve as significant cues for autonomous driving and Advanced Driver Assistance System~(ADAS)~\citep{survey} to keep a car from staying beyond lane markings. Detecting lanes in-the-wild is challenging due to severe occlusion caused by other vehicles, bad weather conditions, ambiguous pavement, and the inherent long and thin property of the lane itself.

\begin{figure}[!t]
\centering
\includegraphics[width=0.45\textwidth]{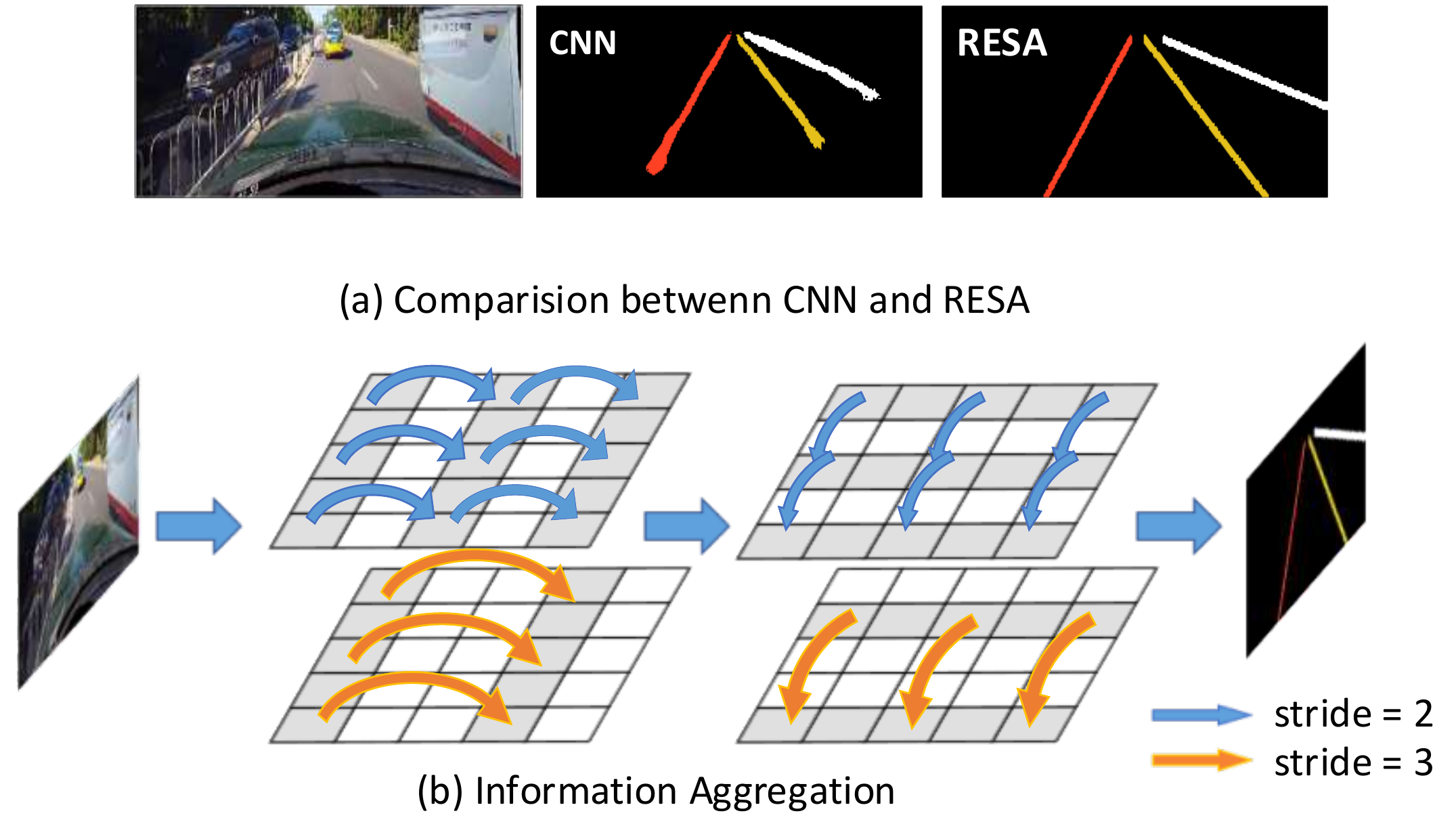}
\caption{
Feature aggregation illustration.
(a) Comparison between CNN semantic segmentation and our method~(RESA). The segmentation method with ordinary CNN suffers from bad performance due to severe occlusion. (b) Illustration of feature aggregation. Spatial lane feature can be enriched, owing to horizontal and vertical feature aggregation in a layer. Thus, RESA can infer lanes even if they are occluded. We add different strides to gather features within different distances, which relieves information loss problem during long-distance propagation.}
\vspace{-5pt}
\label{arch_0}
\end{figure}

Modern algorithms~\citep{inproceedings},~\citep{Bergasa2018},~\citep{Chen2019} typically adopt a pixel-wise prediction formulation, \emph{i.e.}, treat lane detection as a semantic segmentation problem, where each pixel in an image is assigned with a binary label to indicate whether it belongs to a lane. These methods solve the problem with an encoder-decoder framework. They first apply a CNN as the encoder to extract high semantic information into a feature map, then use an up-sampling decoder to recover the feature map to its original size and finally perform a pixel-wise prediction. Due to the thin and long property of lanes, the number of annotated lane pixels is far fewer than background pixels. These methods often struggle to extract subtle lane feature and may ignore the strong shape prior or high relevance between lanes, yielding inferior detection performance. The more challenging case is that the lane may be almost entirely occluded by crowded cars, and we can only conjecture the lane with common sense. Therefore, low-quality feature extracted by ordinary CNN tends to drop subtle lane features. Several methods try to pass spatial information within feature maps, \emph{e.g.}, SCNN~\citep{pan2018spatial}. SCNN typically proposes a spatial convolution to pass information between adjacent rows or columns within feature map. Nevertheless, the sequential information passing operation is time-consuming, which leads to a slow inference speed. Meanwhile, passing information between adjacent rows or columns sequentially takes many iterations, and information may be lost during long-distance propagation.

In this paper, we develop a REcurrent Feature-Shift Aggregator~(RESA) to gather information within feature maps and pass spatial information more directly and efficiently. As shown in Fig.~\ref{arch_0} RESA can aggregate information vertically and horizontally by shifting the sliced feature map recurrently. RESA will first slice the feature map in vertical and horizontal directions, then make each sliced feature receive another sliced feature adjacent to a certain stride. Each pixel is updated simultaneously in several steps, and finally each location can gather information in the whole space. In this way, the information could be propagated between pixels in the feature map.
RESA has three main advantages: 1) RESA passes information in a parallel way, thus reducing time cost significantly. 2) Information will be passed with different strides in RESA. Thus different sliced feature maps can be gathered without information loss during propagation. 3) RESA is simple and flexible to be incorporated into other networks.

Then we propose the Bilateral Up-Sampling Decoder~(BUSD). BUSD has two branches. One is to catch the coarse-grained feature, and the other is to capture the fine-detailed feature. The coarse branch applies the bilinear up-sample directly and produces a blurry image. In contrast, the detailed branch implements up-sample with a transpose convolution and is followed by two non-bottleneck blocks~\citep{romera2017erfnet} to fix fine-detailed loss. Combined with two branches, our decoder can recover the low-resolution feature map into pixel-wise prediction meticulously.

We evaluate our method on two popular lane detection benchmarks, \emph{i.e.}, CULane and Tusimple. Qualitatively, RESA could well preserve the smoothness and continuity of lane detection, as shown in  Fig.~\ref{arch_0}. Furthermore, The experiment results show that RESA achieves state-of-the-art accuracy (75.3 $F_1$-measure on CULane and 96.8\% accuracy on Tusimple).

The main contributions can be summarized as follow:
\begin{itemize}
\item  We propose RESA to aggregate spatial information by shifting sliced feature map recurrently in vertical and horizontal directions. RESA can be easily incorporated into other networks for better performance.
\item The Bilateral Up-Sampling Decoder is further proposed to recover low-resolution feature map meticulously.
\item The network achieves state-of-the-art performance on CULane and Tusimple benchmark. It can serve as a strong baseline to facilitate future research on lane detection.
\end{itemize}

\section{Related Work}

\subsection{Lane Detection}
Lane detection methods can be classified into two classes: traditional methods and deep learning-based methods. Traditional methods try to exploit hand-crafted low-level feature or specialized feature. ~\citet{sun2006hsi} tries to detect lanes in HSI color representation and \citet{yu1997lane} extracts lane boundaries via Hough Transform. These methods require a complex feature selection process and have the weakness of poor scalability due to road scene variations. Recently, deep learning methods have shown superiority in lane detection with the high capacity to learn lane features in the end-to-end manner. \citet{huval2015empirical} are the first to apply the deep learning method in lane detection with CNN. \citet{neven2018towards} propose to cast the lane detection problem as an instance segmentation problem. \citet{philion2019fastdraw} integrates the lane decoding step into the network and draws lanes iteratively without the recurrent neural network. Self-attention distillation~(SAD) is proposed to allow a model to learn from itself and gains substantial improvement without any additional supervision or labels~\citep{hou2019learning}. 

\subsection{Spatial Information Utilization}
There have been some other attempts to utilize spatial information in neural networks.
ION~\citep{bell2016inside} explores the use of spatial Recurrent Neural Networks (RNNs). These RNNs pass spatially varying contextual information both horizontally and vertically across an image. \citet{liang2016semantic} constructs Graph LSTM to provide information propagation route for semantic object parsing. SCNN~\citep{pan2018spatial} proposes to generalize traditional layer-by-layer convolutions to slice-by-slice convolutions within feature maps, thus enabling message passing between pixels across rows and columns in the same layer. SCNN propagates message as residual and makes it easier to train than previous work, but still suffers from expensive computation and information loss during long-distance propagation. RESA is much more computationally efficient than SCNN while gathering information from sliced features with different strides to avoid information loss.

\section{Method}
\label{sec::Method}

\begin{figure*}[!t]
\centering
\includegraphics[width=0.9 \textwidth]{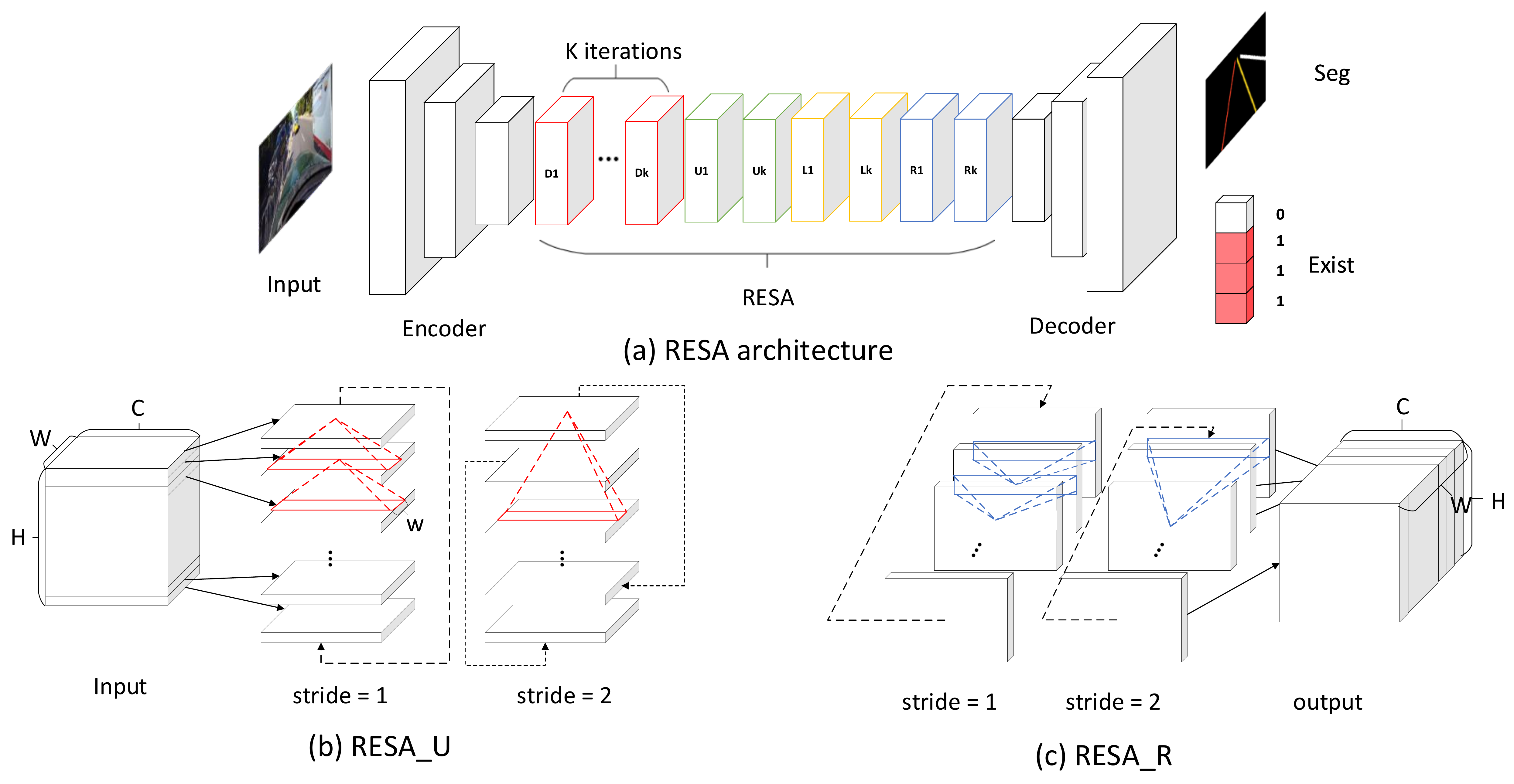}
\caption{Architecture Design. (a) Overall architecture of our model, which is composed by encoder, RESA and decoder. `Dk', `Uk', `Lk', `Rk' denotes ``up-to-down'', ``down-to-up'', ``right-to-left'', and ``left-to-right'' respectively at k-th iteration in RESA. (b) RESA\_U module. In this module, information propagates "down-to-up" with different strides recurrently and simultaneously. (c) RESA\_R module. In this module, information propagates ``left-to-right'' with different strides recurrently and simultaneously.}
\label{arch}
\vspace{-10pt}
\end{figure*}

This section will demonstrate the details of our designed model, including the overall network architecture, RESA, and Bilateral Up-Sampling Decoder.

To take advantage of the strong shape priors of lanes and captures spatial relationships of pixels across rows and columns, we propose a novel RESA module to gather information and enrich the feature map. After inserting RESA into the encoder-decoder framework, our model is constructed with three components: encoder, aggregator, and decoder. We select commonly used backbone like ResNet~\citep{he2016deep}, VGG~\citep{Simonyan2015VeryDC},~\emph{etc} as our encoder to extract preliminary feature from raw image. Then RESA module is applied to aggregate lane feature and get rich feature map. A novel Bilateral Up-Sampling Decoder with coarse-grained branch and fine-detailed branch is proposed to recover lanes smoothly and continuously.

\subsection{Architecture Design}
The overall network architecture is shown in Fig.~\ref{arch}(a). The framework is composed of three components:
\begin{enumerate}
    \item \textit{Encoder:} A commonly used backbone network like VGG, ResNet, and \emph{etc}, is applied as a feature extractor. The size of the raw input image is reduced to 1/8 after passing the encoder. The preliminary feature will be extracted in this stage. 
    \item \textit{RESA:} REcurrent Feature Shift Aggregator~(RESA) is proposed for gathering spatial feature. In every iteration, the sliced feature map will shift recurrently in 4 directions and pass information vertically and horizontally. Finally, RESA needs K iterations to ensure that each location can receive information in the whole feature map.
    \item \textit{Decoder:} Decoder consists of bilateral up-sampling blocks. Each block up-samples two times and finally recover the 1/8 feature map to the original size. Bilateral Up-Sampling Decoder is composed of the coarse-grained branch and fine-detailed branch.
\end{enumerate}

After up-sampled by the decoder, the output feature map is used to predict each lane's existence and probability distribution. A fully-connected layer is followed for existence prediction, and the binary classification will be performed. A pixel-wise prediction will be conducted for lanes probability distribution prediction, which is the same as the semantic segmentation task.

\subsection{Recurrent Feature-Shift Aggregator}
\label{resa_sec}
We propose REcurrent Feature-Shift Aggregator (RESA) to gather spatial information by shifting the sliced feature map horizontally and vertically. Specifically, assume we have a 3-D feature map tensor \textit{X} of size $C\times H\times W$, where $C$, $H$, and $W$ denote the number of channels, rows, and columns, respectively. $X^{k}_{c,i,j}$ means the value of feature map $X$ at \textit{k-th} iteration where $c$, $i$ and $j$ indicate indexes of channel, row and column, respectively. Then the forward computation of RESA is defined as follow:
\begin{align}
	Z^{k}_{c,i,j} &= \sum_{m,n}{F_{m,c,n}\cdot{ X^{k}_{m,(i+s_k) \ mod \ H,j+n-1} }}\label{vertical}, \\
	Z^{k}_{c,i,j} &= \sum_{m,n}{F_{m,c,n}\cdot{ X^{k}_{m,i+n-1,(j+s_k) \ mod \ W} }}\label{horizontal}, \\
	X^{k'}_{c,i,j} &= X^k_{c,i,j} + f(Z^{k}_{c,i,j})\label{fuse_method}, \\
	s_k &= \frac{L}{2^{K-k}}, \quad k=0,1, \cdots, K-1,
\end{align}
where $K = \lfloor log_2 L\rfloor$, $k$ is the iteration number. $L$ in Eq.~\eqref{vertical} and  Eq.~\eqref{horizontal} is $W$ and $H$, respectively. $f$ is a nonlinear activation function as ReLU. The $X$ with superscript $'$ denotes the element that has been updated. $s_k$ is the shift stride in \textit{k-th} iteration. Eq.~\eqref{vertical} and  Eq.~(\ref{horizontal}) show vertical and horizontal information passing formulas. $F$ is a group of 1-d convolution kernel, which size is $N_{in} \times N_{out} \times w$, where $N_{in}$, $N_{out}$ and $w$ denote the number of input channels, the number of output channels and kernel width. Both $N_{in}$ and $N_{out}$ are equal to $C$. $Z$ in Eq.~\eqref{vertical} and  Eq.~\eqref{horizontal} is intermediate results for information passing. Note that feature map \textit{X} is split into $H$ slices in the horizontal direction and $W$ slices in the vertical direction as shown in Fig \ref{arch}(b) and Fig. \ref{arch}(c). We implement recurrent feature-shift information passing simply by index calculation with no other complicated operations. Shift stride $s_k$ is controlled by iteration number $k$, which determines the information passing distance dynamically.

Also, note that the information passing has four directions. We use
``down-to-up'' (shown in Fig.~\ref{arch}(b) RESA\_U), ``up-to-down'' as vertical information aggregator and ``left-to-right'' (Fig.~\ref{arch}(c) RESA\_R), ``right-to-left'' as horizontal information aggregator. The convolution layer weights with the same shift stride are shared across all slices in the same direction.

\begin{figure}[!t]
\centering
\includegraphics[width=0.40\textwidth]{
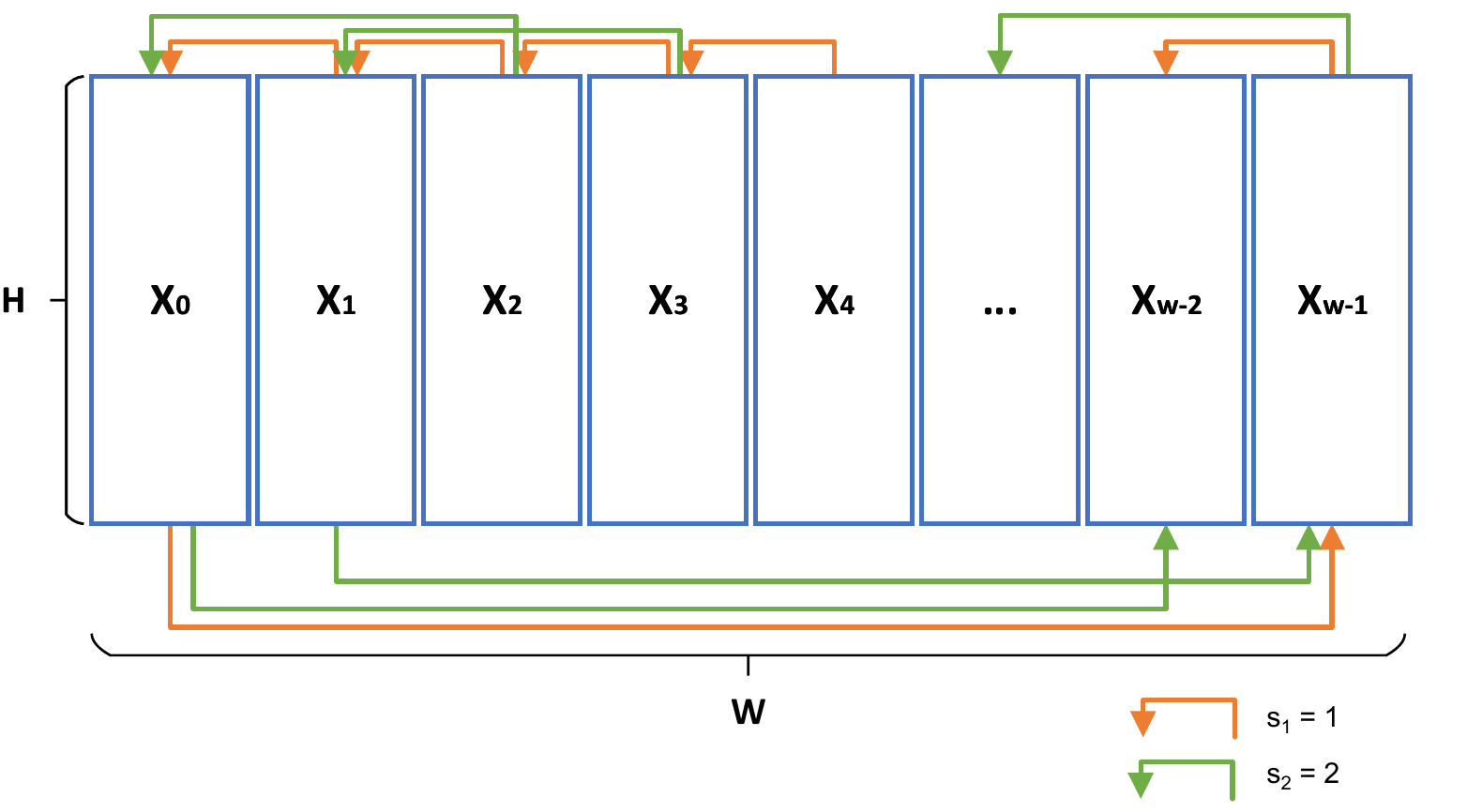}
\caption{Information passing in RESA when $s_1 = 1$ and $s_2 = 2$. $X_0$ can receive information from $X_0$,$X_1$,$X_2$ and $X_3$ only in two iterations.}
\label{message_passing}
\vspace{-10pt}
\end{figure}

We take ``right-to-left'' information passing as a demonstration, and the detail is shown in Fig. \ref{message_passing}. At $k=0$ iteration, $s_1=1$ and $X_{i}$ in each column can receive $X_{i+1}$ shifted feature. Because of recurrently shifting, columns at tail can also receive feature on the other side, \emph{i.e.}, $X_{w-1}$ can receive $X_0$ shifted feature. At $k=1$ iteration, $s_2=2$ and $X_{i}$ in each column can receive $X_{i+2}$ shifted feature. Take $X_0$ as an example, $X_0$ can receive $X_2$ information in the second iteration, considering $X_0$ has received information from $X_1$ while $X_2$ has received information from $X_3$ in the previous iteration, now $X_0$ has received information from $X_0$, $X_1$, $X_2$, and $X_3$ in total only in two iterations. The next iterations are similar to the above procedure. After the all K iterations, each $X_i$ can aggregate information in the whole feature map when iteration $k=K$ finally. 

\subsubsection{Analysis.}
RESA applies feature-shift operation recurrently in 4 directions and enables every location to perceive and aggregate all spatial information in the same feature map. Lane detection is a task that highly relies on surrounding clues. For example, a lane is occluded by several cars, but we can still inference it from other lanes, car direction, road shape, or other visual clues. RESA aggregates feature from other locations to enrich the feature map and helps model to conjecture lanes like humans. The novel and powerful RESA module mainly has three advantages, which are concluded as follows:

\begin{enumerate}
    \item \textit{Computationally efficient.} Traditional information passing methods like Markov Random Field (MRF) or Conditional Random Field (CRF)~\citep{NIPS2011_4296}, where each pixel receives all other pixel information in a fully connected way, always suffer from its intensive and redundant computation. Some methods like SCNN~\citep{pan2018spatial} implement a more effective information passing scheme, \emph{i.e.}, slice-by-slice convolution. However, this RNN-like way still consumes much time as the complexity is increased linearly with the spatial size grows, and the sequential propagation cannot fully utilize computation resources. RESA's complexity is related to spatial size at a $log$ level, and all locations are updated in a parallel way at every iteration. Each location can aggregate information from in the whole feature map in $\lfloor log_2 L\rfloor$ iteration.
    
    \item \textit{Feature information gathered effectively.} The sliced feature information will not only be passed to adjacent slice but also be passed to sliced feature map with different strides, \emph{i.e.}, $s_k = 1, 2, 4, 8, \cdots $. Therefore, each pixel can gather information from the sliced feature map without information loss during propagation. As shown in Fig.~\ref{result}, RESA can get better performance than SCNN since SCNN only passes feature information to adjacent and loses information during propagation.
    
    \item \textit{Easy to be plugged into other networks.} Without bells and whistles, the structure is quite concise. Firstly, the implementation of RESA is simple, which only needs index operation in the feature map. Secondly, RESA doesn't change the shape of the input feature map, which can be treated as a feature enhancement module. An ideal place is after a feature extraction CNN like VGG~\citep{simonyan2014very}, ResNet~\citep{he2016deep}, MobileNet~\citep{Howard2017MobileNetsEC}, and \emph{etc}. Finally, the computational time of RESA can be almost ignored. To sum up, RESA can be plugged into other CNN networks flexibly.  Since RESA is powerful in feature aggregation, scene understanding, and object detection with distinct geometry prior are suitable application scenarios.
\end{enumerate}
    
\subsection{Bilateral Up-Sampling Decoder}
\begin{figure}[!t]
\centering
\includegraphics[width=0.3\textwidth]{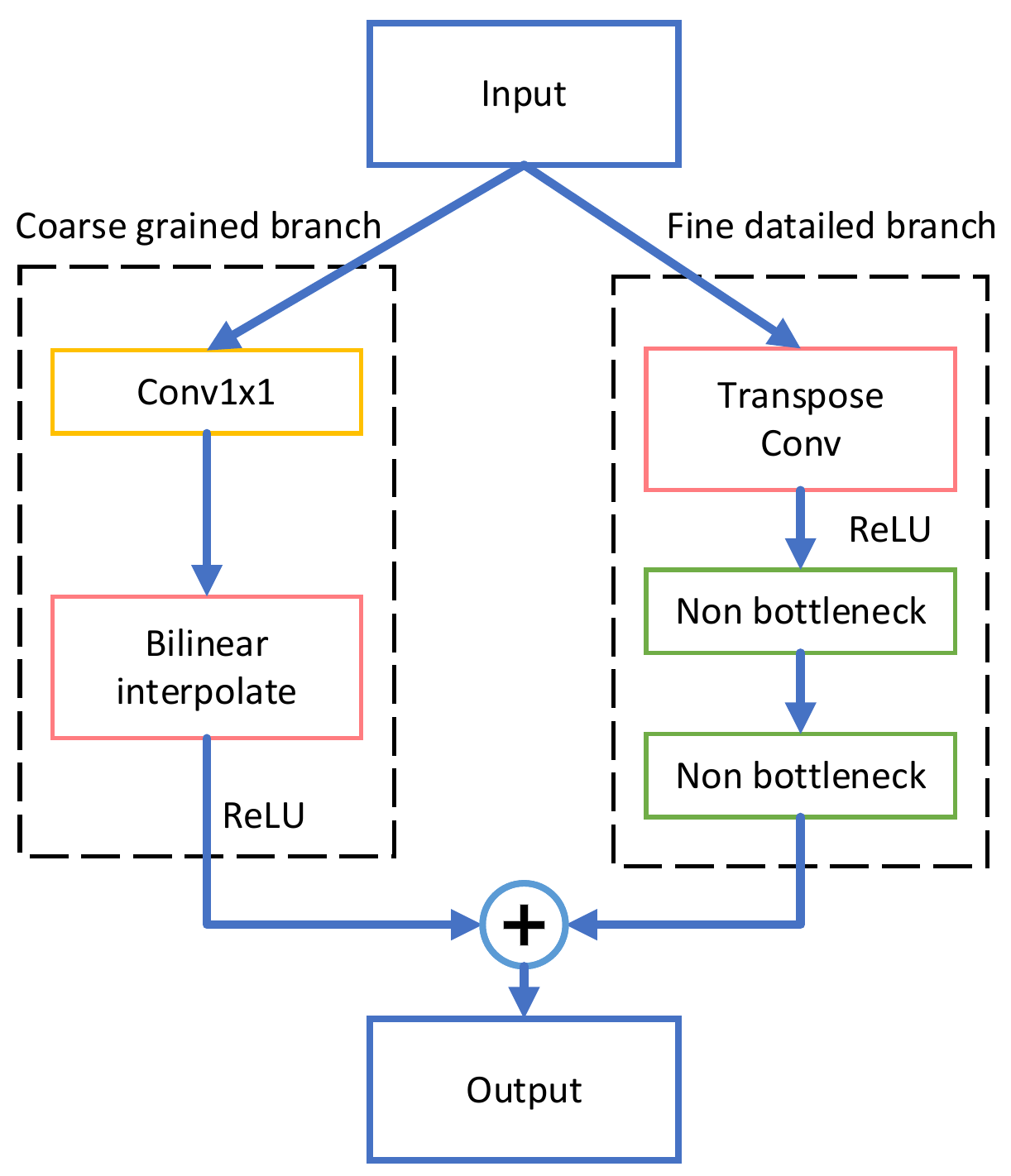}
\caption{Bilateral Up-Sampling Decoder. Decoder up-samples feature map to 2x size. It is composed by coarse grained branch (left) and fine detailed branch (right). Coarse grained branch is used to get a rough up-sampled feature quickly and ignore much detail. Fine detailed branch is used to fine-tune subtle information loss.}
\label{decoder}
\vspace{-12pt}
\end{figure}

The main task of the decoder is up-sampling the feature map to the input resolution. Most decoders utilize the bilinear upsampling procedure to recover the final pixel-wise prediction, which is easy to obtain coarse results but may lose details. Some methods \citep{romera2017erfnet} use stacking convolutional operations and deconvolutional operations to get refined upsampling results. For the motivation above, we combine their advantages and propose Bilateral Up-Sampling Decoder~(BUSD). The decoder is composed of two branches, one is to recover the coarse-grained feature, and the other is to fix fine-detailed loss. The structure is illustrated in Fig.~\ref{decoder}. Input will pass two branches, and 2x up-sampled output with half number of channel will be produced. After passing these stacked decoder blocks, the 1/8 feature map produced by RESA will be recovered to the same size as the input image.

\begin{table*}[]
\centering
\scalebox{1.0}{
\begin{tabular}{llllllll}
\hline
Dataset & \#Frame & Train   & Validation  & Test & Resolution & Scenario Type & \#Lane           \\
\hline
TuSimple   & 6,408 & 3,236 &  358 &  2,782   &  1280 $\times$ 720 & highway &  $\leq$ 5 \\
CULane  & 133,235 & 88,880& 9,675 &  34,680  & 1640 $\times$ 590 & urban, rural and highway & $\leq$ 4       \\
\hline
\end{tabular}
}
\caption{Datasets description.}
\label{datasets}
\end{table*}

\subsubsection{Coarse Grained Branch.}
The coarse-grained branch will output a rough up-sampled feature from the last layer quickly, which may ignore details. A simple and shallow path is designed. We first apply $1\times 1$ convolution to reduce the number of channel by a factor 2 of the input feature map, and a BN~\cite{ioffe2015batch} is followed. A bilinear interpolation is used directly to up-sample the input feature map. At last, the ReLU is performed.

\subsubsection{Fine Detailed branch.}
The fine-detailed branch is used to fine-tune subtle information loss from the coarse-grained branch, and the path is deeper than the other. We use transpose convolution with stride 2 to up-sample feature map and reduce the number of channel by a factor 2 simultaneously. ReLU is followed the up-sampling as the similar design used in the coarse-grained branch. Non-bottleneck block~\citep{romera2017erfnet} consists of four $3\times 1$ and $1\times 3$convolutions with BN and ReLU, which can keep the shape of the feature map and extract information efficiently in a factorized way. We stack two non-bottleneck after the up-sampling operation.

\section{Experiment}

\begin{table*}[]
\centering
\scalebox{0.9}{
\begin{tabular}{c|ccccccc|cc}
    \hline
    Category  & Res50 & Res34-VP &  SCNN    &   Res34-SAD & Res34-Ulrta & PINet & CurveLane-L & RESA-34 &RESA-50\\
    \hline
    Normal       & 87.4 & 90.4  & 90.6 & 89.9 & 90.7 & 90.3 & 90.7 & 91.9 &   \textbf{92.1} \\
    Crowded      & 64.1 & 69.2 & 69.7 & 68.5 & 70.2 & 72.3 & 72.3 & 72.4 &   \textbf{73.1} \\
    Night        & 60.6 & 63.8 & 66.1 & 64.6 & 66.7 & 67.7 & 68.2 & 69.8 &   \textbf{69.9} \\
    No line      & 38.1 & 43.1 & 43.4 & 42.2 & 44.4 & \textbf{49.8} & 49.4 & 46.3 &   47.7 \\
    Shadow       & 60.7 & 62.5 & 66.9 & 67.7 & 69.3 & 68.4 & 70.1 & 72.0 &   \textbf{72.8} \\
    Arrow        & 79.0 & 83.5 & 84.1 & 83.8 & 85.7 & 83.7 & 85.8 & 88.1 &   \textbf{88.3} \\
    Dazzle light & 54.1 & 61.4 & 58.5 & 59.9 & 59.5 & 66.3 & 67.7 & 66.5 &   \textbf{69.2} \\
    Curve        & 59.8 & 64.7 & 64.4 & 66.0 & 69.5 & 65.6 & 68.4 & 68.6 &   \textbf{70.3} \\
    Crossroad    & 2505 & 2141 & 1990 & 1960 & 2037 & \textbf{1427} & 1746 & 1896 &   1503 \\
    \hline 
    Total        & 66.7 & 70.9 & 71.6 & 70.7 & 72.3 & 74.4 & 74.8 & 74.5 &   \textbf{75.3} \\
    \hline
     Runtime~(ms) & - & 26 & 116 & $<$51 & 6 & 40 & - & 22 & 28 \\
        
    \hline
 \end{tabular}}
\caption{Comparison with state-of-the-art results on CULane dataset with IoU threshold = 0.5. For crossroad, only FP are shown. Res50 indicates deeplab~\citep{chen2017deeplab} using resnet50 as backbone.}
\label{culane_main}
\vspace{-8pt}
\end{table*}

\subsection{Dataset}
We conduct experiments on two widely used lane detection benchmark datasets: CULane dataset~\citep{pan2018spatial} and Tusimple ~Lane detection benchmark\footnotemark.
The CULane dataset consists of 55 hours of videos which comprises urban and highway scenarios. It consists of nine different scenarios, including normal, crowd, curve, dazzle night, night, no line, and arrow in the urban area. Tusimple dataset is collected with stable lighting conditions in highways. The details of datasets are showed in Table~\ref{datasets}.
\footnotetext{\url{https://github.com/TuSimple/tusimple-benchmark/}}

\subsubsection{CULane.} For CULane dataset, each lane is treated as a 30-pixel-width line. Intersection-over-union (IoU) is calculated between predictions and ground truth. Predicted lanes whose IoU are larger than a threshold (0.5) are considered as true positives (TP). The F1-measure is taken as the evaluation metric, which is defined as: $F_1 = \frac{2 \times Precision \times Recall}{Precision + Recall}$, where $Precision = \frac{TP}{TP+FP}$ and $Recall = \frac{TP}{TP + FN}$, $FP$ and $FN$ is false positive and false negative respectively.
\subsubsection{Tusimple.} For Tusimple dataset, the evaluation metric is accuracy. It is defined as follow: $accuracy = \frac{\sum_{clip}C_{clip}}{\sum_{clip}S_{clip}}$.
in which $C_{clip}$ is the number of lane points predicted correctly (mismatch distance between prediction and ground truth is within a certain range) and $S_{clip}$ is the total number of ground truth points in each clip. We also evaluate the rate of false positive~(FP) and false negative~(FN) on prediction results.

Following ~\citep{hou2019learning}, we first resize the original images to $288\times800$ for CULane and $368\times 640$ for Tusimple, respectively. We use SGD~\citep{Bottou2010} with momentum 0.9 and weight decay 1e-4 as the optimizer to train our model, and the learning rate is set 2.5e-2 for CULane and 2.0e-2 for Tusimple, respectively. We use warm-up~\citep{Doll2017} strategy in the first 500 batches and then apply polynomial learning rate decay policy~\citep{8929465} with power set to 0.9.

The loss function is the same as SCNN~\citep{pan2018spatial}, which consists of segmentation BCE loss and existence classification CE loss. Considering the imbalanced label between background and lane markings, the segmentation loss of background is multiplied by 0.4. The batch size is set 8 for CULane and 4 for Tusimple, respectively. The total number of training epoch is set 50 for the TuSimple dataset and 12 for the CULane dataset. All models are trained with 4 NVIDIA 2080Ti GPUs~(11G Memory) in Ubuntu. All experiments are implemented with Pytorch1.1.

In our experiments, we use ResNet~\citep{he2016deep} and VGG~\citep{simonyan2014very} as backbone. In ResNet, we add an extra $1\times 1$ convolution to reduce the output channel to 128. The modification of VGG is the same as SCNN.

\subsection{Main Results}
We show the results of our method on two lane detection benchmark datasets and compare it with other popular lane detection methods. For CULane dataset, several popular lane detection methods, including ResNet50~\citep{chen2017deeplab}, Res34-VP~\citep{liu2020heatmap}, SCNN, Res34-SAD~\citep{hou2019learning}, Res34-Ultra~\citep{qin2020ultra}, PINet~\citep{ko2020key}, CurveLane~\citep{xu2020curvelane} are used for comparison. Our RESA adopts ResNet-50 as the backbone, which is marked as RESA-50. The result is shown in Table~\ref{culane_main}. Through the overall design, RESA outperforms all baselines in the CULane dataset and achieves state-of-the-art result. Meanwhile, RESA-50 can achieve 36fps, which denotes our method is efficient in computation and can be used in real-time applications. Moreover, it is observed that our method obtains superior performance in almost all scenarios, which strongly suggests the effectiveness and the generality of RESA.For Tusimple lane detection benchmark, six methods are used for comparison, including ResNet18, ResNet34, ENet~\citep{paszke2016enet}, LaneNet~\citep{wang2018lanenet}, ENet-SAD, and SCNN. We use ResNet-18/34 as the backbone, and they are marked as RESA-18/34. The result is shown in Table~\ref{tusimple_main}. RESA-34 achieves 96.82\% accuracy, which also outperforms the state-of-the-art. We also analyze FP and FN for each method. It is noteworthy that the FP of RESA is far below other algorithms, which means that RESA gains higher precision on the lane detection task and contributes to achieving higher accuracy.

\begin{figure}[!t]
\centering
\includegraphics[width=0.48 \textwidth]{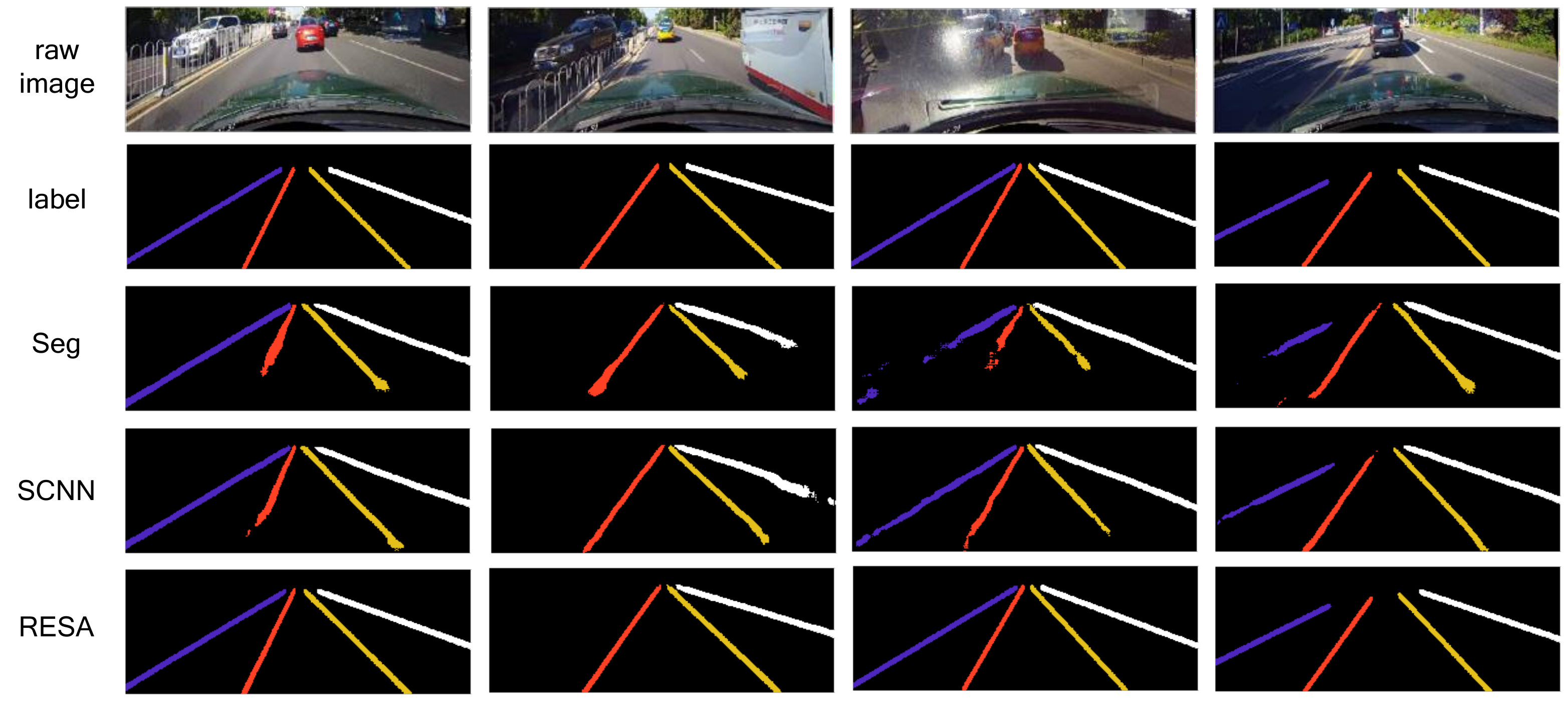}
\caption{Examples results from CULane dataset with segmentation method, SCNN, and RESA.}
\label{result}
\vspace{-10pt}
\end{figure}

To further explain the effectiveness of our method, we show the qualitative results of our algorithm and others in the CULane dataset. As Fig.~\ref{result} shows, segmentation methods cannot preserve lane markings' smoothness and continuity due to severe occlusion. In contrast, SCNN could partially address the problem by passing spatial information and improve the performance, but the result is still unsatisfying. It can be observed that the predictions of SCNN become imprecise at the bottom of the image, where can only be inferred by the surrounding feature. It indicates that information may be lost in SCNN during long-distance propagation. Among these methods, RESA could capture the spatial relationship of the pixel across rows and columns and aggregate information from the sliced feature map with different strides. Therefore, the results of RESA are more robust and contain less noise. This demonstrates that RESA owns much stronger capability to capture structural prior objects than traditional segmentation modules and SCNN.

\begin{table}[!tbh]
\centering
\scalebox{1.}{
\begin{tabular}{c|c|c|c}
\hline
Network & Accuracy & FP & FN \\
\hline
\hline
ResNet-18 & 92.69\% & 0.0948& 0.0822 \\
ResNet-34 &92.84\% & 0.0918& 0.0796 \\
ENet & 93.02\% & 0.0886 & 0.0734 \\
LaneNet & 93.38\% & 0.0780 & 0.0224 \\
ENet-SAD & 96.64\% & 0.0602 & 0.0205 \\
SCNN & 96.53\% & 0.0617 & \textbf{0.0180} \\
\hline
RESA-18~(ours) &96.70\% &0.0395 & 0.0283\\
RESA-34~(ours) & \textbf{96.82\%}&\textbf{0.0363} &0.0248 \\
\hline
\end{tabular}}
\caption{Comparison with state-of-the-art results on Tusimple dataset. ResNet-18/34 indicates deeplab~\citep{chen2017deeplab} using resnet18 and resnet34 as backbone.}
\label{tusimple_main}
\end{table}

\subsection{Ablation Study}
In the \textit{Method} section, we discuss Recurrent Feature-Shift Aggretator~(RESA) and Bilateral Up-Sampling Decoder~(BUSD) and analyze the advantages of each module, respectively. To verify the importance of each proposed component, we make detailed ablation studies in this section.

\subsubsection{Effect of Each Component.}
We first investigated the effectiveness of the Bilateral Up-Sampling Decoder module and RESA module. For the baseline, we select ResNet-34 as the backbone. After being extracted from the backbone, the feature map is up-sampled 8x directly using bilinear interpolation as SCNN does. The output is used as regression and finally gets the probability distribution of each lane. 

To make comparison, we replace bilinear interpolation with Bilateral Up-Sampling Decoder and then insert RESA between backbone and decoder step by step. We summarize the performance of each module in Table~\ref{tab:Module ablation}. As we can see, both modules can strongly improve lane detection performance, which proves the capabilities of proposed modules.

\begin{table}[!tbh]
	\centering
	\begin{tabular}{cccc}
			\toprule
			Baseline & BUSD & RESA & F1 \\ \midrule
			\checkmark& & & 65.1 \\
			& \checkmark & & 68.6~(+3.5) \\
			& & \checkmark & 74.3~(+9.2) \\
			& \checkmark &\checkmark & \textbf{74.5~(+9.4)} \\ \bottomrule
	\end{tabular}
	\caption{Experiments of the proposed modules on CULane dataset with ResNet-34 backbone. Baseline stands for 8x up-sampling directly after backbone.}
	\label{tab:Module ablation}
	\vspace{-10pt}
\end{table}

\subsubsection{Effectiveness of Feature Aggregation.}
In this section, we investigate the effect of the direction in RESA.
As we add more directions in RESA, we can get higher F1-measure. 
The result is shown in Table.~\ref{direction}.
Furthermore, we study the feature aggregation method in Eq.~\ref{fuse_method}.
We replace the addition operator by maximum operator. The result shows that the maximum operator has similar performance as addition operator.

\begin{table}[!tbh]
\centering
\scalebox{1.0}{
\begin{tabular}{c|c|c|c}
\hline
Method  & Precision & Recall & F1-measure \\
\hline
\hline
RESA\_U & 74.2 & 70.6 & 72.4 \\
RESA\_UR & 75.1 & 72.3	&  73.7 \\
RESA\_LRDU & 76.7 & 72.3 & 74.4  \\
RESA\_DULR & 76.1 & 72.9 & \textbf{74.5} \\
RESA\_DULR$^{\dagger}$ &76.0 & 72.8 & 74.4\\
\hline
\end{tabular}}
\caption{Effectiveness of feature aggregation of RESA on
CULane dataset with ResNet-34 backbone.
$\dagger$ means maximum feature aggregation method.
}
\label{direction}
\vspace{-10pt}
\end{table}

\subsubsection{Iteration In RESA.}
In this section, we explore the effect of different iterations in RESA. Theoretically, as the iteration increases, each slice of feature map can aggregate more information, which contributes to obtain better performance. To illustrate more iterations can bring up better performance, we make comparison between different iterations, \emph{i.e.}, $iteration = 1,\cdots,5$. As shown in Table~\ref{iter}, the performance will be better as the iteration increases. However, more iterations lead to more computational time cost. It is a trade-off between performance and computational resources. To strike a balance between them, we select $iteration = 4$ as our final choice.

\begin{table}[!tbh]
\centering
\scalebox{1.0}{
\begin{tabular}{c|c|c|c}
\hline
Iter  & Precision & Recall & F1-measure \\
\hline
\hline
1 & 74.7 & 71.7&	73.2 \\
2 &74.4	 &72.4 & 	73.4\\
3 &74.8	& 72.5 &	73.6   \\
4 &76.1	& \textbf{72.9} &	\textbf{74.5}  \\
5 &\textbf{76.9}	&72.1 & \textbf{74.5} \\
\hline
\end{tabular}}
\caption{The performance of the model by using different iterations on CULane dataset with ResNet-34 backbone.}
\label{iter}
\vspace{-10pt}
\end{table}


\subsubsection{Compare RESA with SCNN.}
SCNN~\citep{pan2018spatial} has shown message passing scheme could improve the lane detection performance but extra more parameters could merely bring about little improvement. Thus, we compare the RESA with SCNN to verify the effectiveness of our method.  We try to add RESA and SCNN with different backbones~(\emph{e.g.} ResNet, VGG). We conduct experiment to compare the performance with SCNN. The experiment results are shown in Table~\ref{resa_scnn}. The result shows that RESA outperforms SCNN and brings significant improvement. 

\begin{table}[!tbh]
\centering
\scalebox{1.0}{
\begin{tabular}{c|c|c|c}
\hline
Method  & Precision & Recall & F1-measure \\
\hline
\hline
VGG16 & 62.2 &	60.3&	61.2  \\
VGG16 + SCNN & 72.4 & 72.1 & 72.3 \\
VGG16 + RESA &\textbf{74.1} &\textbf{72.5} & \textbf{73.3}\\
\hline
ResNet34 & 66.2& 64.2& 65.1 \\
ResNet34 + SCNN &73.9 & 71.5&	72.7 \\
ResNet34 + RESA & \textbf{76.1} & \textbf{72.9}& \textbf{74.5}  \\
\hline
\end{tabular}}
\caption{The comparison between SCNN and RESA trained using VGG16 and ResNet34 as backbone.}
\label{resa_scnn}
\vspace{-5pt}
\end{table}

%
%

\subsubsection{Computational Efficiency.}
We also conduct experiment to compare the running time of our method with LSTM, SCNN. The running time of these methods are recorded with the average time for 1000 runs. We use different convolution kernel widths (7, 9, 11) to compare the efficiency. SCNN propagates information in a sequential way, \emph{i.e.}, a slice does not pass information to the next slice until it has received information from previous slice. Thus, this kind of message passing requires much computational cost due to sequential computing. In contrast, our RESA passes information in a parallel way. As shown in Table~\ref{relation}, RESA is around 10 times faster than SCNN with the same kernel width, which makes it promising to apply our method to real-time applications.

\begin{table}[!tbh]
\centering
\scalebox{0.85}{
\begin{tabular}{c|c|ccc|ccc}
\hline
Method  & LSTM & \multicolumn{3}{|c}{SCNN} &  \multicolumn{3}{|c}{RESA} \\
\hline
\hline
Kernel width & - & 7 & 9 & 11  & 7 & 9 & 11 \\
Runtime~(ms) & 108.0 & 43.5 & 44.0 & 44.6 & \textbf{3.8} & 4.0 & 4.4 \\
\hline
\end{tabular}}
\caption{Runtime of LSTM, SCNN, and RESA. The iteration in RESA is 4.}
\label{relation}
\end{table}
\vspace{-5pt}

\section{Conclusion}
In this paper, we propose two components tailored for lane detection: Recurrent Feature-Shift Aggretator~(RESA) and Bilateral Up-Sampling Decoder~(BUSD). RESA takes the advantage of strong shape priors of lanes and captures spatial relationships of pixels across rows and columns. It shifts sliced feature map recurrently in vertical and horizontal directions and enables each pixel to gather global information. Besides, it can be plugged into other networks easily. The Bilateral Up-Sampling Decoder is proposed to combine coarse grained feature and fine detailed feature in up-sampling stage. Our method is evaluated on two popular lane detection benchmark datasets, \emph{i.e.}, Tusimple and CULane and achieves the state-of-the-art performance.

\section*{Acknowledgments}
This work was supported in part by The National Key Research and Development Program of China (Grant Nos: 2018AAA0101400), in part by The National Nature Science Foundation of China (Grant Nos: 62036009, U1909203, 61936006, 61973271).

\bibliography{aaai}

\end{document}